\documentclass[twoside,leqno,twocolumn]{article}

\usepackage[letterpaper]{geometry}

\usepackage{ltexpprt}

\usepackage{amsmath,amssymb,amsfonts}
\usepackage{algorithmic}
\usepackage{graphicx}
\usepackage{textcomp}
\usepackage{xcolor}

\usepackage{algorithm}
\usepackage{subfigure}

\usepackage{diagbox}
\usepackage{threeparttable}
\usepackage{multirow}
\usepackage{color}
\usepackage{amsmath}
\usepackage{url}

\usepackage[utf8]{inputenc} 
\usepackage[T1]{fontenc}    
\usepackage{booktabs}       
\usepackage{amsfonts}       
\usepackage{nicefrac}       
\usepackage{microtype}      
\usepackage{amsmath}
\usepackage{multirow}
\usepackage{caption}

\usepackage{placeins}

\makeatletter
\renewcommand{\@thesubfigure}{\hskip\subfiglabelskip}
\makeatother

\begin{document}


\title{Disentangled Dynamic Graph Deep Generation}


\author{Wenbin Zhang\thanks{University of Maryland, Baltimore County, MD 21250, USA. Email: wenbinzhang@umbc.edu}
	\newline
	\and Liming Zhang~$^\dag$
	\and Dieter Pfoser \thanks{George Mason university, VA 22030, USA. }
	\and Liang Zhao$^1$\thanks{Emory University, GA 30322, USA. Email: liang.zhao@emory.edu		
\newline $^1$corresponding author}}
\date{}

\maketitle


\fancyfoot[R]{\scriptsize{Copyright \textcopyright\ 2021 by SIAM\\
Unauthorized reproduction of this article is prohibited}}






\begin{abstract}
	
Deep generative models for graphs have exhibited promising performance in ever-increasing domains such as design of molecules (i.e, graph of atoms) and structure prediction of proteins (i.e., graph of amino acids). Existing work typically focuses on static rather than dynamic graphs, which are actually very important in the applications such as protein folding, molecule reactions, and human mobility. Extending existing deep generative models from static to dynamic graphs is a challenging task, which requires to handle the factorization of static and dynamic characteristics as well as mutual interactions among node and edge patterns. Here, this paper proposes a novel framework of factorized deep generative models to achieve interpretable dynamic graph generation. Various generative models are proposed to characterize conditional independence among node, edge, static, and dynamic factors. Then, variational optimization strategies as well as dynamic graph decoders are proposed based on newly designed factorized variational autoencoders and recurrent graph deconvolutions. Extensive experiments on multiple datasets demonstrate the effectiveness of the proposed models.

\end{abstract}



\section{Introduction}
Graphs are ubiquitous data structures to capture connections (i.e., edges) among individual units (i.e., nodes), such as social networks, proteins (network of amino acids), and molecules (network of atoms). One central problem in data mining and machine learning for graphs is the gap between the discrete graph topological information and continuous numerical vectors preferred by mathematical models~\cite{guo2020systematic}. This directly leads to two major directions on graph research: 1) graph representation learning~\cite{kazemi2019relational}, which aims at encoding graph structural information into (low-dimensional) vector space,
and 2) graph generation~\cite{guo2020systematic}, which reversely aims at constructing a graph-structured data from low-dimensional space containing the graph generation rules or distribution. Graph generation is an important domain with long history that attracts extensive models such as random graphs~\cite{robins2001random}, small world models~\cite{aksoy2019generative}, scale-free graphs~\cite{alam2019novel}, and stochastic block models~\cite{louail2015uncovering}. They typically rely on the network generation principles predefined by human heuristics and prior knowledge, which effectively abstract the high-dimension problems down to manageable scale. Such methods usually fit well towards the properties that the predefined principles are tailored for, but usually cannot do well for the others~\cite{radford2015unsupervised}. Unfortunately, the underlying principle of many critical real-world problems is still unknown, such as generating new molecule under desired biophysical properties and simulating brain functional connectivity.

In recent years, the success of deep generative models in image and text generation~\cite{xie2017neural,goodfellow2014generative} has been extended to graph data applications such as molecule design~\cite{simonovsky2018graphvae} and protein structure prediction~\cite{alquraishi2019alphafold}. They typically leverage frameworks such as Variatioal Autoencoder (VAE)~\cite{hsu2018scalable} and Generative Adversarial Nets (GAN)~\cite{bojchevski2018netgan}, which rely on highly expressive deep architectures to map high-dimensional graph-structured data into latent low-dimensional space, where the latent data points just follow simple distributions. However, existing works on deep generative models of graphs typically focus on static graphs instead of dynamic graphs. Although research on dynamic networks is a trending deep learning topic, it mostly focuses on representation learning tasks, including node/graph embedding~\cite{radford2015unsupervised,nguyen2018continuous}, node/graph classification~\cite{bhagat2011node,zhang2018end}, link prediction~\cite{liben2007link,zhang2020deep}. However, the domain of deep generative models for dynamic graph generation has not been well explored~\cite{guo2020systematic,kazemi2019relational}.

Extending the current deep generative models from static graphs to dynamic graphs are very important to many critical domains, such as modeling protein folding process~\cite{simonovsky2018graphvae}, human mobility networks~\cite{jiang2017activity}, and dynamic functional connectivity process in human brains~\cite{borsboom2017network}. Despite its enormous importance, several major challenges hinder this tasks from being easily achieved by existing techniques, including: \textbf{1) Difficulty in jointly modeling node dynamics and edge dynamics.} Nodes could involve continuous values in their attributes while edges could involve graph topology information that is discrete. Some node dynamic patterns and edge dynamic patterns could be coupled with each other while the other of their patterns may be independent. It is difficult to build a generic, expressive models that can automatically learn and stratify all such patterns from data. \textbf{2) Difficulty in characterizing dynamic and static components.} A dynamic graph is typically a mixture of both time-evolving patterns as well as stationary patterns. For example, during protein folding, its backbone connection tends to be stable as a chain structure while other connections such as hydrogen bonding could vary over time. How deep generative models can identify, disentangle, and model static and dynamic patterns in dynamic graphs is an open problem. \textbf{3) Lack of graph decoders for dynamic node and edge joint generations.} Deep generative models require decoding latent representations back to graph domain, which itself is a nascent, open, and promising domain. How to further extend it to dynamic graph decoding is deemed even more interesting and challenging, which requires to further consider the temporal dependency~\cite{kazemi2019relational}.

To address the above challenges, this paper proposes a novel framework of Disentangled deep generative models for interpretable Dynamic Graph Generation (D2G2). New generative models are proposed to characterize conditional independence among node, edge, static, and dynamic factors. Then, varational optimization strategies are proposed based on newly designed factorized variational autoencoders. Third, novel dynamic node-edge co-decoders are proposed based on recurrent and graph deconvolutions. The contributions of this paper can be summarized as follows:

\begin{itemize}
\item A generic framework of deep generative models for interpretable dynamic graph generation.
\item New deep graph models that jointly characterize node, edge, static, and dynamic factors. 
\item New dynamic graph factorized varational autoencoders for inferring the  established graph models.
\item New dynamic graph decoder architecture to jointly decode time-evolving nodes and edges.
\item Extensive experiments and case studies.
\end{itemize}


\section{Related Work}
\label{sec: relatedWork}

\textbf{Temporal graph generation.} Research on temporal graph generation has mainly focused on extending graph theory into temporal graphs. A plethora of approaches, including randomized reference models~\cite{karsai2011small}, stochastic block models~\cite{louail2015uncovering} and models based on temporal random walk~\cite{starnini2012random} to name a few, have been proposed with applications in numerous domains. Such methods
usually fit well towards the properties that the predefined principles are tailored for, but usually cannot
do well for the others~\cite{radford2015unsupervised}.

\noindent\textbf{Representation learning on dynamic graphs.} Fast increase attentions have been attracted in this research direction to encode graph's structural properties into nodes, edges, or the whole graph embeddings. Representative works include temporal sequence embedding~\cite{xiao2017wasserstein}, graph sequence embedding~\cite{xu2018graph2seq} and more recently continuous-time embedding~\cite{nguyen2018continuous}. A comprehensive recent literature survey covering this research effort is provided in~\cite{kazemi2019relational}.   

\noindent\textbf{Deep generative models for the graphs.}
Deep generative neural networks have achieved the state of the art results for graph generation in various domains, such as molecule design~\cite{simonovsky2018graphvae} and cyber-network synthesis~\cite{guo2018deep}. Most of the existing works in this thread are based on variational autoencoders (VAE)~\cite{kingma2013auto} and generative adversarial networks (GAN)~\cite{goodfellow2014generative}. For example, GraphVAE~\cite{simonovsky2018graphvae} utilizes the VAE model to learn the representation of graphs but also node features; while GraphRNN~\cite{you2018graphrnn} decomposes the graph generation into a sequence of node and edge formations that can be learned by autoregressive models. NetGAN~\cite{bojchevski2018netgan} and its extensions~\cite{guo2020systematic} are trained with the GAN algorithm to generate synthetic random walks while discriminates synthetic walks from real random walks sampled from a real graph. The critical limitation of existing works is that they majorly focus on static networks and cannot disentangle the dynamic and static patterns for nodes and edges.

\noindent\textbf{Disentangled generative models.} Disentangled representation learning aims at learning distinguishable underlying representations that responsible for different variations in the data. Such representations have been utilized to improve generalizability~\cite{alemi2016deep} and interpretability~\cite{dai2018syntax}. A surge of models have been proposed for extending the training objective to enhance disentanglement in various deep generative model architecture such as varational autoencoders and generative adversarial nets~\cite{chen2018isolating,hsu2017unsupervised}. However, most existing works investigate in the field of image representation learning and the disentanglement of the latent factors behind a graph has not been well explored. Our approach fills this gap and further addresses dynamic graph generation.

\section{Disentangled Dynamic Graph Deep Generation (D2G2)}
\subsection{Notations and Problem Formulation}
\begin{figure*}
	\centering
	\includegraphics[width=0.9\textwidth, clip]{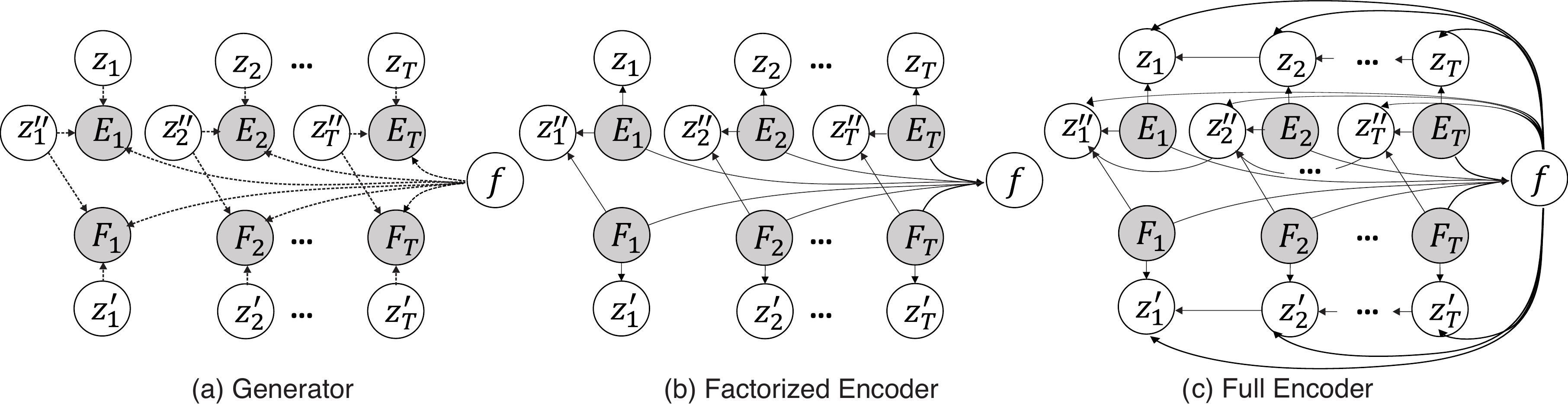}\vspace{-0.2cm}
	\caption{Graphical illustration of the proposed models. (a) The Bayesian network of the proposed probabilistic distribution of dynamic graphs. (b) The approximate inference model of the posterior of the proposed model, with conditional independence assumption among latent variables across different time points. (c) The alternative approximate inference model of the posterior of the proposed model, with dependence among time-variant and invariant variables and dependence across different time points.}
	\label{fig: model}
\end{figure*}
Formally, for different time intervals $0,1,\cdots,T$, a dynamic graph $G := \{G_0, G_1, \cdots, G_T\}$ is described by its topological and attributed information, where for each time step $t\in T$ we have the graph snapshot $G_t=(E_t,F_t)$ where the adjacency matrix is denoted as $E_t \in \{0, 1\}^{n\times n}$ and node feature vector is represented as $F_t \in {\Bbb R}^{n\times c}$. Here $n$ is the maximum number of nodes and each node has $c$ features. If there is a connection between nodes $i$ and $j$ then the element $[E_t]_{i,j}=1$; otherwise, we have $[E_t]_{i,j}=0$. Note that we sometimes use the notations $E=\{E_t\}_{t=0}^T$ and $F=\{F_t\}_{t=0}^T$ for simplicity. In this paper, we focus on \emph{factorized deep generative models for dynamic graphs} that can learn the distribution $p(G|\Theta)$ of dynamic graphs $G$ generated from latent semantic variables $\Theta=\{f,z,z',z''\}$ that characterize the generative process for time-invariant graph patterns (by $f$) and time-variant graph patterns, which can be further factorized into edge-exclusive patterns (by $z$), node-exclusive patterns (by $z'$), and node-edge joint patterns (by $z''$).

\subsection{Factorized Bayesian Models of Dynamic Graphs}
To achieve the above goal, we formulate the proposed generative model as a Bayesian network, as shown in Figure \ref{fig: model}(a). Here as mentioned above, a dynamic graph $G_T$ is a sequence of snapshots of edge-node pairs $\{(E_1,F_1),(E_2,F_2),\cdots,(E_T,F_T)\}$, which are collectively controlled by a time-independent latent variable $f$ to absorb the time-invariant patterns related to the intrinsic nature of the graphs that do not change over time. To characterize the time-variant patterns, additional latent variables $z$, $z'$, and $z''$ are introduced. Here $z$ is to model the dynamics only related to edges. For example, in \emph{dynamic functional connectivity} of a human brain, the co-activation patterns among different brain regions can change quickly over time. $z'$ is to model the dynamics merely pertain to nodes. For example, in traffic network where nodes are road segments and edges are their connections, only the node attributes such as traffic flow changes but their connections (i.e., road network) are unchanged. In addition, there are also dynamics that simultaneously involve both nodes and edges such as the process of \emph{homophily and influence} in social networks. Here we leverage the variable $z''$ to account for it. In the following, we use the subscript $t$ to denote any variable specific to the time $t$. In all, the generative process can be introduced as follows:
\begin{align}
\label{eq:model}
    &p(E,F|z,z',z'',f)=p(E|z,z'',f)p(F|z',z'',f)\\\nonumber =&\prod\nolimits_t p(E_t|z_t,z''_t,f)\cdot p(F_t|z'_t,z''_t,f)
\end{align}

\subsection{Variational Model Inference}
%
%
Since the true posterior $p(z,z',z'',f|E,F)$ of the proposed generative model is intractable to infer, we proposed to solve it based on variational inference where the posterior needs to be approximated by another distribution $q(z,z',z'',f|E,F)$. So the goal is to minimize the Kullback–Leibler (KL) divergence between the true and approximate posteriors. Here, we will first introduce the approximate posteriors and then introduce the objective functions for model inference based on them.

\noindent\textbf{Inference models.} We propose two different inference models based on two different assumptions on the conditional independence among variables, namely a \emph{factorized inference model} and a \emph{full inference model}.

\noindent\emph{Factorized inference model}. As shown in Figure \ref{fig: model}(b), the inference model can be formulated as a factorization for the latent variables with mutual conditional independence, as below:
\begin{align}\nonumber
    q(z,z',z'',f|E,F)=q(z|E) q(z'|F) q(z''|E,F) q(f|E,F)
\end{align}
where we assume the conditional independence among all the variables $z,z',z''$ and $F$.

Here we further decompose the model along the temporal dimension, by assuming the temporally conditional independence among the variable across different time points, by the following equations:
\begin{align}
    q(z|E)=\prod\nolimits_t q(z_t|E_t),\ q(z'|F)=\prod\nolimits_t q(z'_t|F_t), \\\nonumber
    \ q(z''|E, F)=\prod\nolimits_t q(z''_t|E_t,F_t)
\end{align}
where the latent variables for different time points are conditionally independent from each other given the observations for the corresponding time points.

\noindent\emph{Full inference model}. We could also weaken our assumption of conditional independence among time-variant and time-invariant variables, by an alternative inference model shown in Figure \ref{fig: model}(c). Specifically, we have:
\begin{align}
    &q(z,z',z'',f|E,F)\\\nonumber=&q(z|E,f)\cdot q(z'|F,f)\cdot q(z''|E,F,f)\cdot q(f|E,F)
\end{align}
where we can see by assuming $z,z'$ and $z''$ to depend on $f$, we allow that the time-variant patterns of graphs are dependent on intrinsic static properties of the graphs. 

We further decompose the model along the temporal dimension, by allowing temporal dependence among the variables of different time points as follows:
\begin{align}\nonumber
    q(z|E)=\prod\nolimits_t q(z_t|z_{<t},E_t),\ q(z'|F)=\prod\nolimits_t q(z'_t|z'_{<t},F_t), \\
    \ q(z''|E,F)=\prod\nolimits_t q(z''_t|z''_{<t},E_t,F_t)
\end{align}
where the latent variables for different time points are dependent on their previous time points as well as the observations for the corresponding time points.

\noindent\textbf{Objective functions}
Given the above-defined inference models, the inference of the posterior of the proposed generative models in Equation \eqref{eq:model} requires to minimize the $KL(q(z,z',z'',f|E,F)||p(z,z',z'',f|E,F))$, where $KL(\cdot)$ denotes KL divergence. It is equivalent to maximizing the following evidence lower bound (ELBO) of our model:
\small
\begin{equation}\nonumber
\begin{aligned}
\label{equ:marginal}
\max_{\theta,\phi} \mathbb{E}_{\Theta\sim q_{\phi}(\Theta | E, F)} \big[ \log( p_{\theta}({E}, {F} | \Theta) ) \big]\!  -\! KL(q_{\phi}(\Theta |E,F) || p(\Theta))
\end{aligned}
\end{equation}\normalsize
where $\Theta=\{z,z',z'',f\}$ is the set of all the variables while $\theta$ and $\phi$ explicitly denote the parameters of the distributions corresponding to our generator and inference model, respectively. The prior distribution $p(\Theta)=p(z,z',z'',f)=p(z)p(z')p(z'')p(f)$, where each variable follows an isotropic Gaussian distribution. Recall that we have proposed two inference models which are the two possible options for instantiating $q_{\phi}(\Theta|E,F)$, namely factorized inference model and full inference model. When employing different inference model we have the corresponding objective function. First, for factorized inference model, we have:
\small
\begin{equation}\nonumber
\begin{aligned}
\label{equ:marginal}
\max_{\theta,\phi} \mathbb{E}_{\Theta\sim q_{\phi}(\Theta | E, F)} \big[ \log( p_{\theta}({E}| z,z'',f)p_{\theta}(F|z',z'',f) ) \big]  - \\\sum\nolimits_t \big[KL(q_{\phi}(z_t|E_t)||p(z_t)) + KL(q_{\phi}(z'_t|F_t)||p(z'_t))+\\KL(q_{\phi}(z''_t|E_t,F_t)||p(z''_t))+KL(q_{\phi}(f|E_t,F_t)||p(f))\big]
\end{aligned}
\end{equation}\normalsize
Alternatively, for the full inference model, we have:
\small
\begin{equation}
\begin{aligned}
\label{equ:marginal}
\max_{\theta,\phi} \mathbb{E}_{\Theta\sim q_{\phi}(\Theta | E, F)} \big[ \log( p_{\theta}({E}| z,z'',f)p_{\theta}(F|z,z'',f) ) \big]  - \\\nonumber\sum\nolimits_t\!\! \big[KL(q_{\phi}(z_t|z_{<t},\!E_t,\!f)||p(z_t)) \! + \! KL(q_{\phi}(z'_t|z'_{<t},\!F_t,\!f)||p(z'_t))\\\nonumber+KL(q_{\phi}(z''_t|z''_{<t},E_t,F_t,f)||p(z''_t))+KL(q_{\phi}(f|E_t,F_t)||p(f))\big]
\end{aligned}
\end{equation}\normalsize

\subsection{Model Architecture} The model inference objectives proposed in the previous section are equivalent to learning the parameters $\theta$ and $\phi$ for the distributions of generator and inference models. However, due to the extremely complicated distribution patterns for graph structured data, it is prohibitively difficult to predefine any prescribed simple distribution and hence a highly expressive model that can flexibly learn the (unknown) distribution is preferred. Inspired by the recent success in learning complicated distributions for complex data such as images and texts, we leverage graph deep learning models to fit the underlying distributions of generator and inference models, parameterized by neural network parameters $\theta$ and $\phi$, respectively.

\noindent\textbf{Architecture of the generator:} First, for edge generation, we use a dense layer with the inputs including $z,z''$ and $f$. Then, we utilize a graph deconvolutional network \cite{guo2018deep}  to generate a matrix of edge probabilities on each time snapshot. For node generation, we use another dense layer which takes the input that is the concatenation of $z', z''$, and $f$, and output each node features on each time snapshot.






\noindent\textbf{Architecture of full inference model.} First, we utilize a Graph Convolutional Network \cite{kipf2016semi} to produce a representation vector ${a_t}$ for each snapshot graph: ${a_t} = GCN(E_t)$. Raw node attributes are also converted to a feature vector through a dense layer ${b_t} = Dense(F_t)$ first. 
To encode $f$, $q_{\phi}(f| E, F)$ is modeled with a neural network function $h_{\phi}(E, F) \mapsto f$. In details, we concatenate latent representation of topology $a_t$ and attribute representation $b_t$ and feed into a Bi-directional LSTM $BiLSTM$ as inputs. Then, both the last forward output $m_{T}$ and the first backward output ${\bar m_T}$ are concatenated to a single vector. This vector is passed to another dense layer to get $f$. The function $h_{\phi}(E, F) \mapsto f$ is decomposed as follow:
\begin{align}
    {a_t} = GCN(E_t), {b_t} = Dense(F_t)\\\nonumber
    m_{t}, {\bar m_t} = BiLSTM(a_t || b_t),\ f = Dense(m_{T} || {\bar m_T})
\end{align}
where $||$ is the concatenation operation. With similar way, we leverage another three Bi-directional LSTMs to encode $z,z'$ and $z''$. For the details of them, please refer to our supplementary material.




\noindent\textbf{Architecture of factorized inference model.} The encoding structure for $f$ is the same as the above. The only difference focuses on the encoding of $z_t$, $z'_t$, and $z''_t$. Different from the above full inference model, we simply use three multi-layer perception models to generate them, using the inputs $E_t$, $F_t$, and jointly $E_t,F_t$, respectively. No bi-directional LSTM is needed here.

\subsection{Complexity Analysis}

The computational efficiency is an essential prerequisite for large-scale graph generation. Most of the existing graph generation methods require   
$\mathcal{O}(n^3)$, e.g., GraphVAE, or even $\mathcal{O}(n^4)$, e.g., Li et al., time in general, thus limiting their applications in generating small graphs. The overall time complexity of the proposed D2G2 is $\mathcal{O}(cTn^2)$ and is therefore capable of generating modest scale attributed graphs with hundreds or thousands of nodes.

\section{Experiments}
In this section, the performance of our model is evaluated using several synthetic and real-world datasets against the state-of-the-art, on various aspects including quantitative, qualitative and efficiency analyses. The experiments were performed on a 64-bit machine with a 10-core processor (i9, 3.3GHz), 64GB memory with GTX 1080Ti GPU.
\subsection{Datasets}
We evaluate our proposed D2G2 on both real and synthetic benchmark datasets with graph sizes $n$ ranging from 8 to 2500 and diverse characteristics.

\textbf{Protein.} The dynamic folding process of a protein peptide is simulated with a sequence AGAAAAGA. For graph learning, this can be considered as a graph of 8 nodes with node attributes $(x,y,z)$ corresponding to 3D coordination of the $C_\alpha$ atom of each amino acid, producing 300 temporal graphs with a sequence length of 100.     

\textbf{Authentication.} 97 users' authentication graphs~\cite{kent2016cyber}. Each user's graphs are generated by authentication activities on their accessible 27 computers or servers, i.e., $n= 27$, in an enterprise computer network during a 485h period. All times are normalized in the range of [0, 1]. 

\textbf{Metro.} Metro graph data captured by farecard records from the Washington D.C. metro system which reflects million of users' trips records from May 2016 to July 2016. There are 91 stations as graph nodes and each day is treated as a temporal graph sample.

\textbf{Synthetic datasets.} We also consider three synthetic datasets with increasing complexity from scale-free random graphs~\cite{barabasi1999mean}. Existing works use it as static graph, we append a continuous-time value to generated edge in each constructing step to simulate it as dynamic graphs, resulting 3 synthetic datasets with number of nodes $n= 100, 500, 2500$, respectively.
\begin{figure*}[!htb]\vspace{-0.7cm}
	\centering
	\subfigure[(a) Ground truth\vspace{-0.4cm}]{\includegraphics[width=0.25\textwidth]{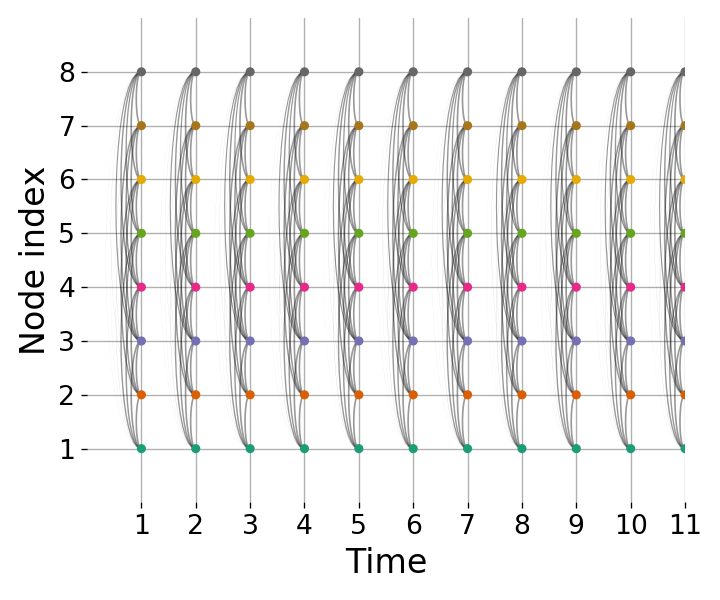}\vspace{-0.4cm}}
	\subfigure[(b) D2G2\vspace{-0.4cm}]{\includegraphics[width=0.25\textwidth]{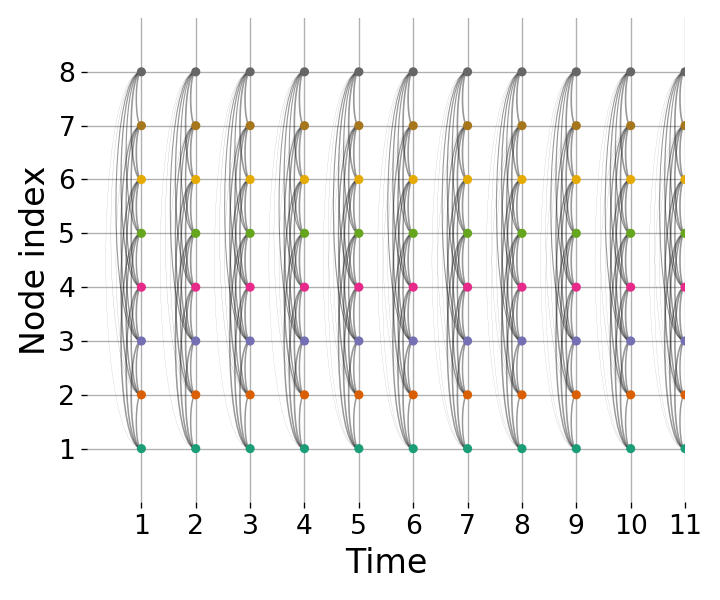}\vspace{-0.4cm}}
	\subfigure[(c) dsbm\vspace{-0.4cm}]{\includegraphics[width=0.25\textwidth]{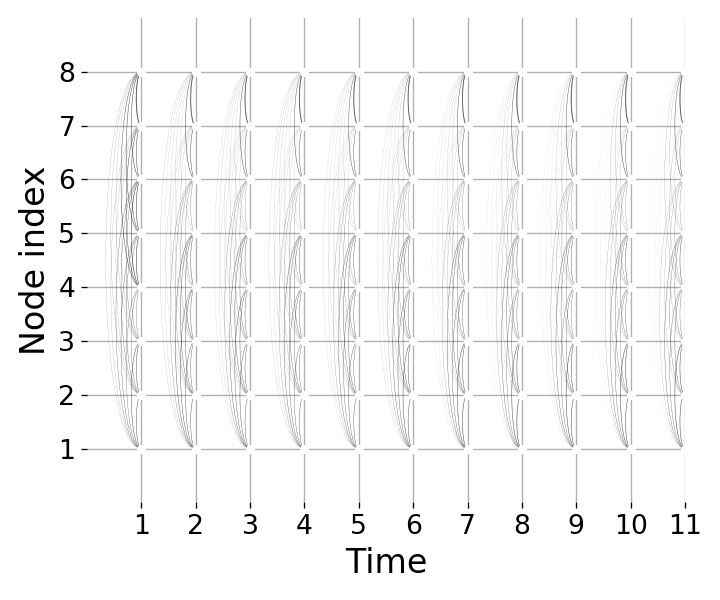}\vspace{-0.4cm}}
	\subfigure[(d) graphrnn\vspace{-0.4cm}]{\vspace{-0.4cm}\includegraphics[width=0.24\textwidth]{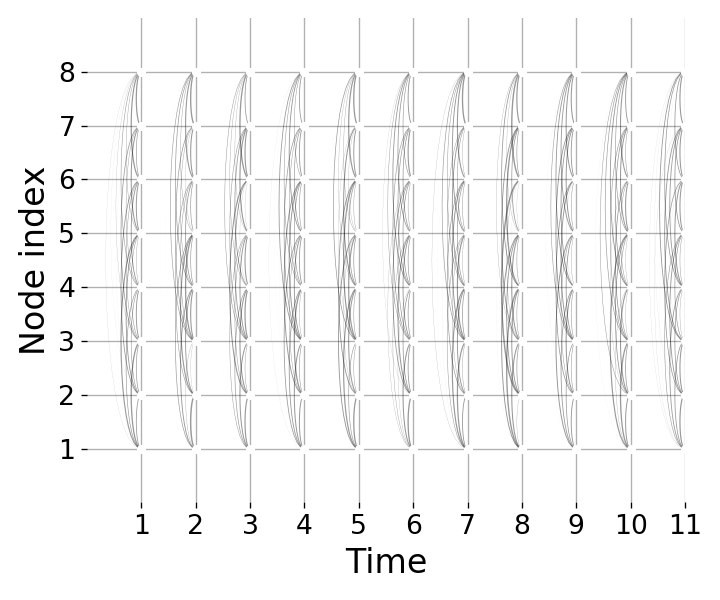}\vspace{-0.4cm}}
	\subfigure[(e) graphvae\vspace{-0.4cm}]{\vspace{-0.4cm}\includegraphics[width=0.24\textwidth]{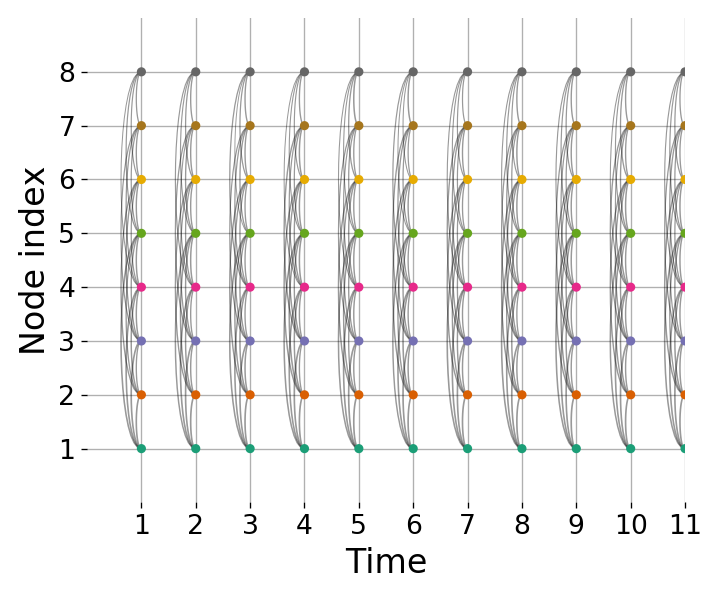}\vspace{-0.4cm}}
	\subfigure[(f) netgan\vspace{-0.4cm}]{\vspace{-0.4cm}\includegraphics[width=0.24\textwidth]{./figures/all/netgan.png}\vspace{-0.4cm}}
	\vspace{-0.4cm}
	\caption{Visualization of graphs from Protein dataset and generated by all methods. The color of nodes represents the normalized value of a node attribute.\vspace{-0.4cm}}
	\label{fig: topology}
\end{figure*}

\subsection{Experimental Setup}
We compare our method with several state-of-the-art deep graph generative models based on a set of metrics measuring the similarity between real and generated graphs in topology and node features. 

\begin{table}[h]\scriptsize
    \captionsetup{font=footnotesize}
    \caption{Comparison of D2G2 to baseline models using MMD metrics with the best performance marked in boldface.\vspace{-0.2cm}}
	\scriptsize
	\label{table: mmd}
	\centering
	\scalebox{0.7}{
	\begin{tabular}{ m{.35in} m{.5in} m{.45in}  m{.45in}  m{.45in}  m{.45in}  m{.45in} m{.45in}}
		\toprule
		Dataset & \diagbox[width=.65in]{Method}{Metrics} & Betweenness Centrality & Broadcast Centrality & Burstiness Centrality & Nodes' Temporal Correlation &  Receive Centrality & Temporal Correlation \\
		\midrule
		\multirow{4}{*}{Protein} 
		& GraphRNN 
		& 0.8160 & 0.0224 & 1.25e-04 & 0.9166 & \textbf{0.0126} & 0.0822 \\
		& NetGAN 
		& 0.9305 & 0.2907 & 9.43e-05 & 1.6151 & 0.2244 & 0.3663 \\
		& GraphVAE 
		& 0.8036 & 0.0295 & 8.20e-05 & 0.9220 & 0.0145 & 0.0821\\
		& DSBM 
		& 0.4169 & 0.5787 & 1.06e-05 & 1.5632 & 0.7479 & 0.4446 \\
		\cline{2-8}
		& \textbf{D2G2} 
		& \textbf{0.3042} & 0.0099 & \textbf{6.75e-07} & \textbf{0.0017} & 0.0174 & \textbf{1.19e-04} \\
		& \textbf{D2G2\_full} 
		& \textbf{0.3042} & \textbf{0.0098} & \textbf{6.75e-07} & \textbf{0.0017} & 0.0173 & \textbf{1.19e-04} \\
		\hline
		\multirow{4}{*}{Auth.} 
		& GraphRNN 
		& 0.2281 & 0.0390 & 0.0160 & 0.2899 & 0.1128 & 5.35e-04 \\
		& NetGAN 
		& 0.1418 & 0.6548 & 0.0184 & 0.4874 & 0.6714 & 0.0376 \\
		& GraphVAE 
		& 0.2084 & 0.1040 & 0.0141  & 0.3216 & 0.0979 & \textbf{9.91e-04} \\
		& DSBM 
		& 0.3255 & \textbf{0.0020} & 0.0035 & 0.2697 & \textbf{0.0053} & 0.0052\\
		\cline{2-8}
		& \textbf{D2G2} 
		& \textbf{0.0017} & 0.0379 & \textbf{2.86e-05} & \textbf{0.2339} & 0.0365 & 0.0549 \\
		& \textbf{D2G2\_full} 
		& \textbf{0.0017} & 0.0379 & \textbf{2.86e-05} & \textbf{0.2339} & 0.0365 & 0.0549 \\
		\hline
		\multirow{4}{*}{Metro} 
		& GraphRNN 
		& NaN & 0.0526 & 1.28e-04 & 0.6854 & 5.93e-05 & 0.0023 \\
		& NetGAN 
		& 1.0 & 4.52e-04 & 0.0146  & 1.9655 & 2.07e-04 & 0.5425\\
		& GraphVAE 
		& NaN & 0.07405 & 6.00e-05  & 0.6742 & 0.0282 & \textbf{3.73e-04} \\
		& DSBM 
		& NaN & \textbf{4.43e-07} & 6.54e-06 & 0.3088 & \textbf{4.60e-07} & 0.0030 \\
		\cline{2-8}
		& \textbf{D2G2}
		& \textbf{0.0878} & 1.59e-06 & \textbf{1.46e-06} & \textbf{0.1189} & 3.50e-06 & 0.0018 \\
		& \textbf{D2G2\_full}
		& 0.0879 & 1.59e-06 & \textbf{1.46e-06} & \textbf{0.1189} & 3.50e-06 & 0.0018 \\
		\hline
		\multirow{4}{*}{100} 
		& GraphRNN & 0.9567 & \textbf{0.1658} & 0.3790  & {0.0011} & 0.3023 & 1.81e-06 \\
		& NetGAN & 0.6497 & 0.7058 & 0.0092  & 0.0014 & 0.2878 & 7.31e-06 \\
		& GraphVAE & 0.9567 & 0.2167 & 0.4138   & 0.0011 & 0.3539 & 1.81e-06 \\
		& DSBM     & \textbf{0.0020} & 0.4016 & 0.0526 &  0.0183 & \textbf{0.1317} & 1.33e-04 \\
		\cline{2-8}
		& \textbf{D2G2}
		& 0.0204 & 0.1874 & \textbf{1.98e-05}  & \textbf{9.12e-05} & 0.2357 & \textbf{1.69e-06} \\
		& \textbf{D2G2\_full}
		& 0.0203 & 0.1873 & \textbf{1.98e-05}  & \textbf{9.12e-05} & 0.2358 & \textbf{1.69e-06} \\
		\hline
		\multirow{3}{*}{500} 
		& GraphRNN & 0.7912 & 0.1556 & 0.1621 & {0.0049} & 0.4241 & \textbf{1.75e-07} \\
		& NetGAN & 0.7928 & 0.3253 & {0.0415} & {0.0049} & \textbf{0.0948} & \textbf{1.75e-07}\\
		& GraphVAE & 0.7928 & 0.0871 & 0.1921  & {0.0049} & 0.2858 & \textbf{1.75e-07} \\
		\cline{2-8}
		& \textbf{D2G2}
		& \textbf{0.0015} & \textbf{0.0124} & \textbf{4.19e-05}  & \textbf{6.09e-04} & 0.1904 & 7.85e-07\\
		& \textbf{D2G2\_full}
		& \textbf{0.0015} & \textbf{0.0124} & \textbf{4.19e-05}  & \textbf{6.09e-04} & 0.1904 & 7.85e-07\\
		\hline
		\multirow{2}{*}{2500} 
		& GraphRNN & 0.8802 & 1.0239 & \textbf{9.65e-07}  & 0.0044 & 1.2410 & \textbf{1.00e-08} \\
		& NetGAN & 0.8801 & 0.1200 & 0.0169  & 0.0044 & \textbf{0.0965} & \textbf{1.00e-08} \\\cline{2-8}
		& \textbf{D2G2}
		& \textbf{0.0002} & \textbf{0.0056} & {7.25e-05}  & \textbf{0.0005} & 0.1487 & 8.18e-05 \\
		& \textbf{D2G2\_full}
		& \textbf{0.0002} & \textbf{0.0056} & {7.25e-05}  & \textbf{0.0005} & 0.1487 & 8.18e-05 \\
		\hline
	\end{tabular}}
\end{table}\vspace{-0.4cm}

\begin{table}[!htb]\vspace{-0.2cm}
	\centering
	\captionsetup{font=footnotesize}
	\scriptsize
	\caption{Comparison of D2G2 to baseline models based on node attribute generation with the best performance marked in boldface.\vspace{-0.2cm}}
	\label{table: nodeEva}
	\setlength{\tabcolsep}{20pt}
	\scalebox{0.7}{
	\begin{tabular}{ccccc}
		\toprule
		Dataset & \diagbox[width=.95in]{Method}{Metrics} & MSE & R2 & PCC \\
		\midrule
		\multirow{2}{*}{Protein} 
		& {\tiny GraphVAE} & 0.0080 & 0.83 & 0.72 \\
		\cline{2-5}
		& \textbf{D2G2} & \textbf{0.0040} & \textbf{0.97} & 0.97  \\
		& \textbf{D2G2\_full} & \textbf{0.0040} & 0.96 & \textbf{0.98}  \\
		\cline{1-5}
		\multirow{2}{*}{Auth.} 
		& {\tiny GraphVAE} & 0.0036 & 0.95 & 0.87  \\
		\cline{2-5}
		& \textbf{D2G2} & \textbf{0.0024} & \textbf{0.99} & \textbf{0.93}  \\
		& \textbf{D2G2\_full} & \textbf{0.0024} & \textbf{0.99} & \textbf{0.93}  \\
		\cline{1-5}
		\multirow{2}{*}{Metro} 
		& {\tiny GraphVAE} & 0.0006 & 0.92 & 0.82  \\
		\cline{2-5}
		& \textbf{D2G2} & 0.0004 & \textbf{0.95} & 0.86  \\
		& \textbf{D2G2\_full} & \textbf{0.0003} & \textbf{0.95} & \textbf{0.87}  \\
		\cline{1-5}
		\multirow{2}{*}{100} 
		& {\tiny GraphVAE} & 2.13 & \textbf{0.86} & 0.89  \\
		\cline{2-5}
		& \textbf{D2G2} & \textbf{1.97} & 0.76 & \textbf{0.91}  \\
		& \textbf{D2G2\_full} & \textbf{1.97} & 0.76 & \textbf{0.91}  \\
		\cline{1-5}
		\multirow{2}{*}{500} 
		& {\tiny GraphVAE} & 45.90 & 0.45 & 0.39  \\
		\cline{2-5}
		& \textbf{D2G2} & 43.77 & \textbf{0.52} & \textbf{0.47}  \\
		& \textbf{D2G2\_full} & \textbf{42.77} & 0.51 & \textbf{0.47}  \\
		\cline{1-5}
		\cline{1-5}
		\multirow{2}{*}{2500} 
		& {\tiny GraphVAE} & 15.56 & 0.75 & 0.78  \\
		\cline{2-5}
		& \textbf{D2G2} & 1.49 & \textbf{0.89} & \textbf{0.81}  \\
		& \textbf{D2G2\_full} & \textbf{1.48} & \textbf{0.89} & \textbf{0.81}  \\
		\cline{1-5}
	\end{tabular}}\vspace{-0.2cm}
\end{table}

\textbf{Comparison Methods.} 1) GraphVAE~\cite{simonovsky2018graphvae} is the pioneering variational autoencoder based graph generation method; 2) NetGAN~\cite{bojchevski2018netgan} trains the graph generative model with the GAN algorithm; 3) GraphRNN~\cite{you2018graphrnn} is the recent generative model for graphs based on sequential node and edge generation; 4) Dynamic-Stochastic-Blocks-Model (DSBM)~\cite{xu2014dynamic}, which extends the stochastic blockmodel for static networks to the dynamic setting. Among them, GraphVAE is the only existing method that achieves graph disentanglement, i.e., learning the topology of graphs but also node features. 5) Disentangled Dynamic Graph Deep Generation (D2G2), our proposed models (factorized and full) in this paper whose parameter setting is elaborated in the supplementary material.

\textbf{Evaluation metrics.} Various maximum mean discrepancy (MMD) metrics~\cite{gretton2007kernel}, including betweenness centrality, broadcast centrality, burstiness centrality, node temporal correlation, receive centrality and temporal correlation are utilized to evaluate the performance of our models and other baselines in terms of topology simulation. The lower these MMD values the merrier a generative model. Additional measures including mean squared error (MSE), coefficiency of determination score (R2) and Pearson correlation coefficient (PCC) are also evaluated for node attribute evaluation to shed full light on the similarity between the real target and generated graphs.

\subsection{Generating Attributed Temporal Graphs}

We present both quantitative and qualitative experiment results that demonstrate the effectiveness and efficiency of D2G2 in generating attributed temporal graphs. 

\begin{figure*}[!htb]
	\centering
	\subfigure{\includegraphics[width=0.24\textwidth, height=0.02\textheight]{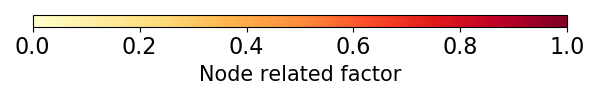}}\vspace{-0.4cm}
	\subfigure[(a) Generated graphs when varying $f$ with $z$, $z'$ and $z''$ fixed.]{
		\begin{minipage}{\textwidth}
			\subfigure{\includegraphics[width=0.24\textwidth]{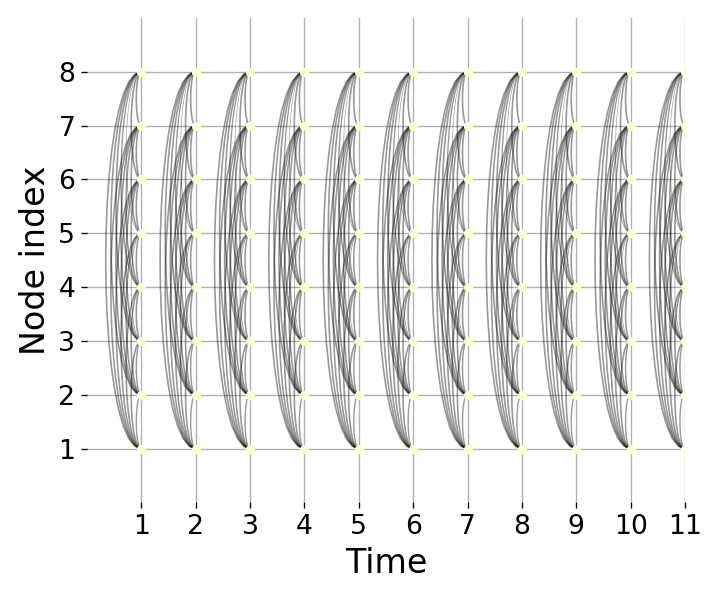}\vspace{-0.4cm}}
			\subfigure{\includegraphics[width=0.24\textwidth]{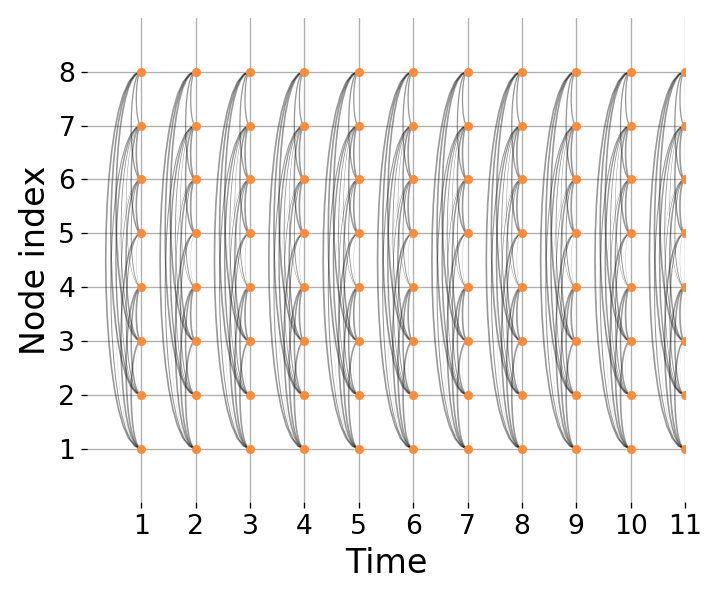}\vspace{-0.4cm}} 
			\subfigure{\includegraphics[width=0.24\textwidth]{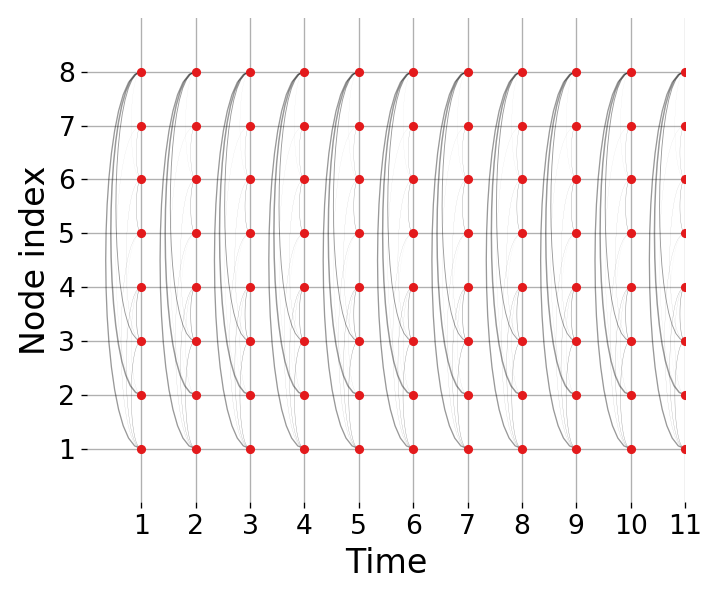}\vspace{-0.4cm}} 
			\subfigure{\includegraphics[width=0.24\textwidth]{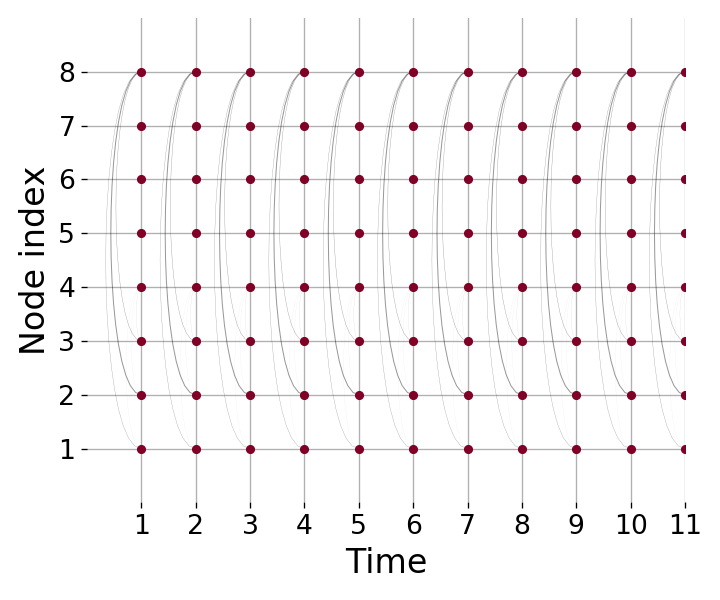}\vspace{-0.4cm}}
			\label{fig: fix_f}
		\end{minipage}
	}

	\vspace{-0.4cm}\subfigure[(b) Generated graphs when varying $z$ with $f$, $z'$ and $z''$ fixed.]{
		\begin{minipage}{\textwidth}
			\subfigure{\includegraphics[width=0.24\textwidth]{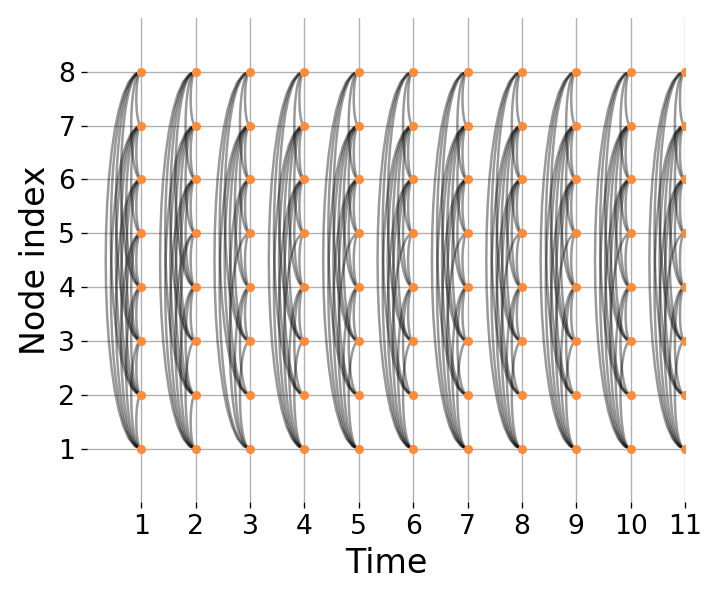}\vspace{-0.4cm}}
			\subfigure{\includegraphics[width=0.24\textwidth]{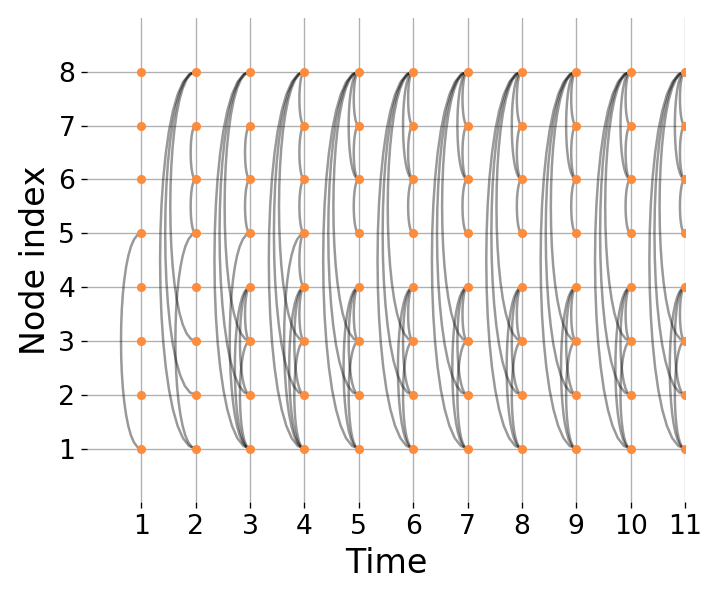}\vspace{-0.4cm}} 
			\subfigure{\includegraphics[width=0.24\textwidth]{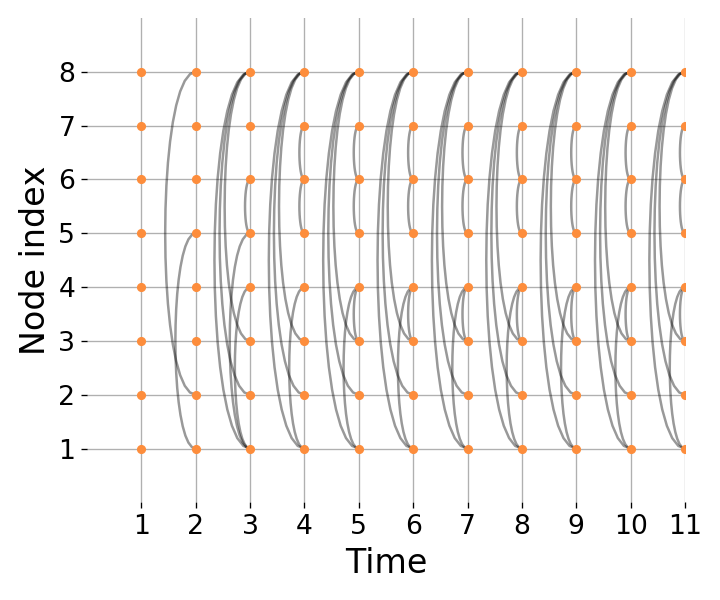}\vspace{-0.4cm}}
			\subfigure{\includegraphics[width=0.24\textwidth]{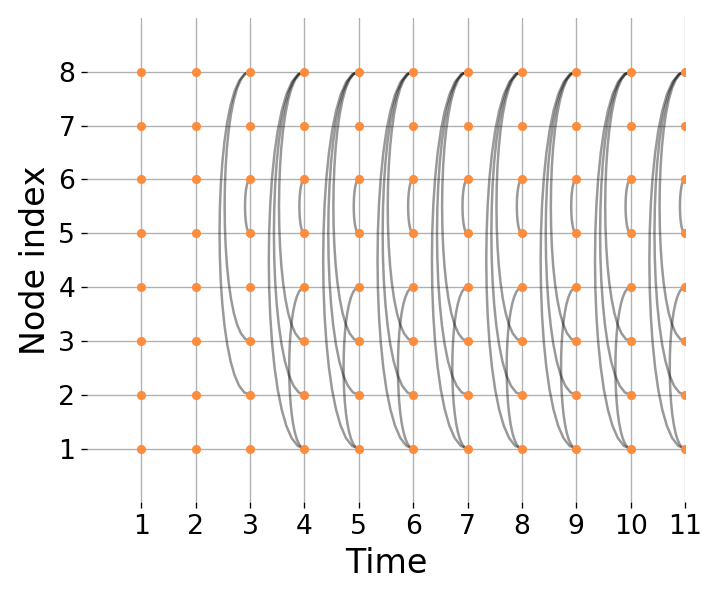}\vspace{-0.4cm}}
		\end{minipage}
	}
	
	\vspace{-0.4cm}\subfigure[(c) Generated graphs when varying $z'$ with $f$, $z$ and $z''$ fixed.\vspace{-0.4cm}]{
	\begin{minipage}{\textwidth}
		\subfigure{\includegraphics[width=0.24\textwidth]{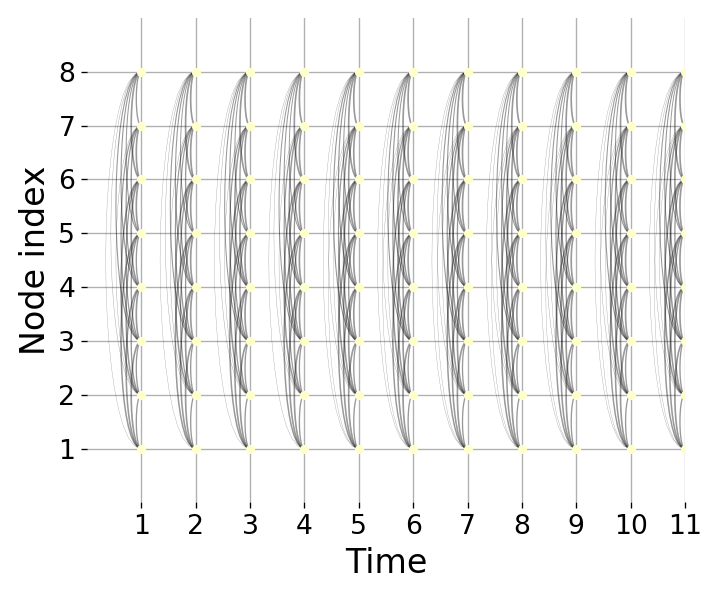}\vspace{-0.4cm}}
		\subfigure{\includegraphics[width=0.24\textwidth]{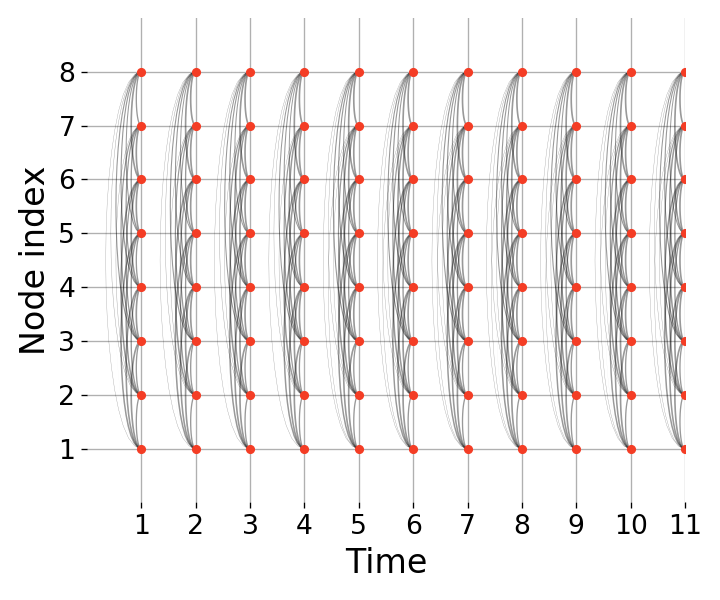}\vspace{-0.4cm}} 
		\subfigure{\includegraphics[width=0.24\textwidth]{./figures/fix_z_prime/protein_fix_z_prime_3.png}\vspace{-0.4cm}}
		\subfigure{\includegraphics[width=0.24\textwidth]{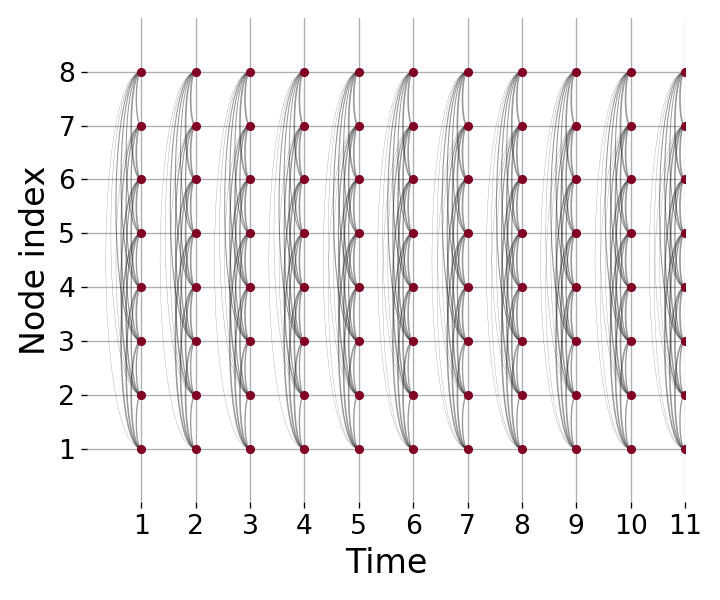}\vspace{-0.4cm}}
	\end{minipage}
	}

			\vspace{-0.4cm}\subfigure[\vspace{-0.4cm}(d) Generated graphs when varying $z''$ with $f$, $z$ and $z'$ fixed.\vspace{-0.4cm}]{
		\begin{minipage}{\textwidth}
			\subfigure{\includegraphics[width=0.24\textwidth]{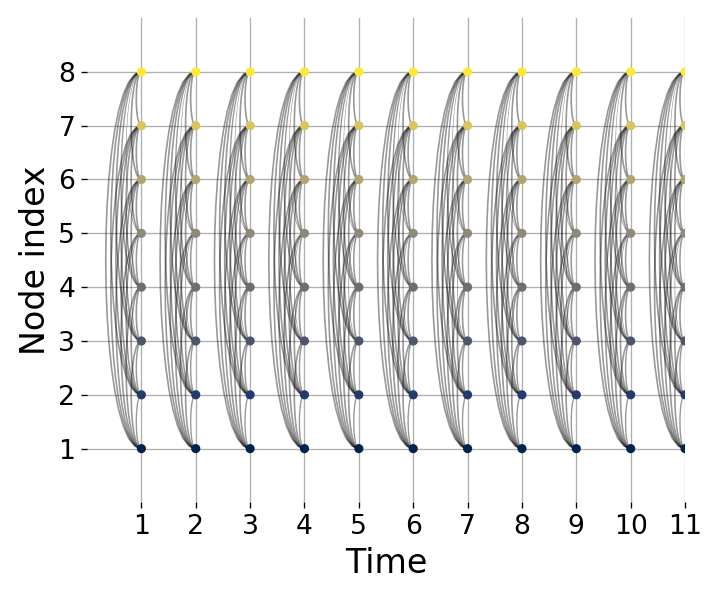}\vspace{-0.4cm}}
			\subfigure{\includegraphics[width=0.24\textwidth]{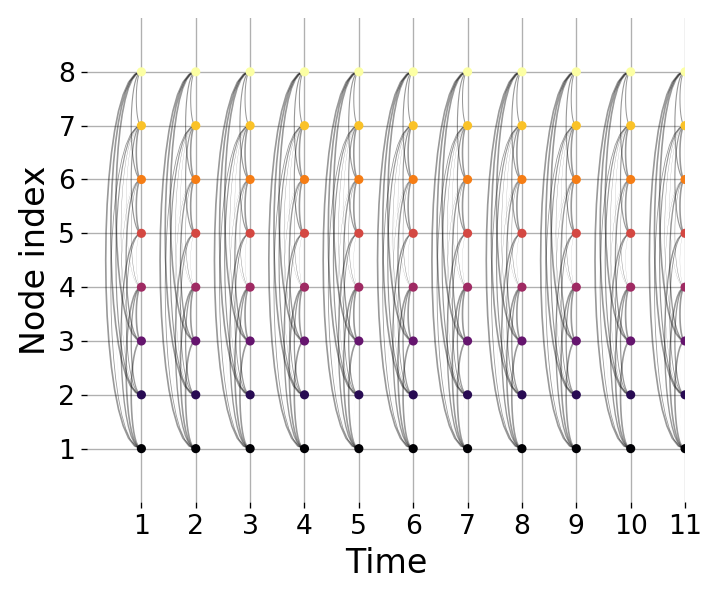}\vspace{-0.4cm}} 
			\subfigure{\includegraphics[width=0.24\textwidth]{./figures/fix_z_doublePrime/protein_fix_z_doublePrime_3.png}\vspace{-0.4cm}}
			\subfigure{\includegraphics[width=0.24\textwidth]{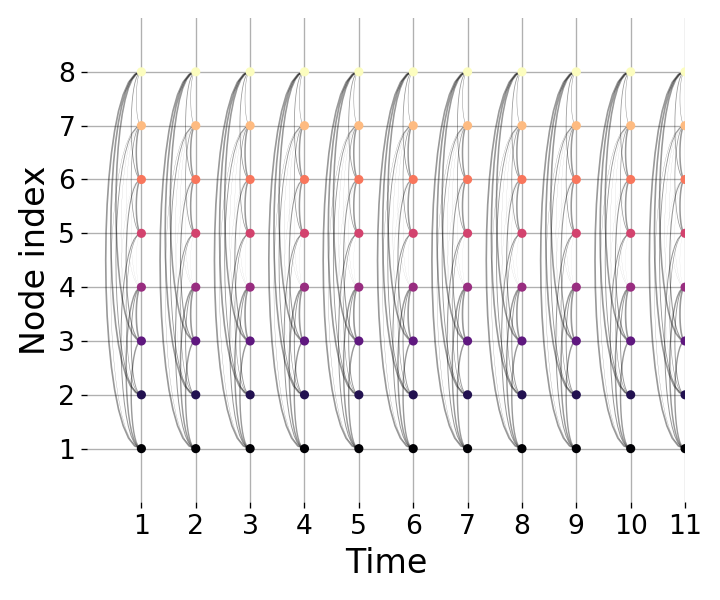}\vspace{-0.4cm}}
		\end{minipage}
	}\vspace{-0.4cm}

	\caption{Visualization of Protein graphs generation controlled by respective disentangling factors.\vspace{-0.3cm}}
	\label{fig: caseStudy}
\end{figure*}

\textbf{Evaluation with graph statistics.} A set of topology related MMD metrics and node attribute evaluations are tested on different models and the results are summarized in Tables~\ref{table: mmd} and~\ref{table: nodeEva}, respectively. 

As one can see, D2G2 based models consistently achieve superior performance on most of metrics across all datasets in both node and attribute generation, and are the only methods to handle both properly. On the other hand, although some baseline models perform well on certain specific datasets, they cannot do well for other types of input graphs. In addition, while methods such as DSBM does not scale well and can only generate smaller graphs, D2G2 enjoys the merit of handling graphs with increasing complexity. In terms of node attributes, GraphVAE is the only method capable of handling node attribute prediction, the proposed D2G2 outperforms GraphVAE significantly and can be as high as 88\% on related statistics while scaling to thousand level graphs where GraphVAE fails. The superior performance in both graph topology and node attributes generation of D2G2 verifies the necessity of its theoretically disentangling design, which captures the underlying graph dynamics very well. Note that in the conducted experiments, factorized and full encoder networks produce almost identical results, presumably because respective factors are truly independent. We therefore only report results on factorized network in the following discussions.

\textbf{Temporal graph visualization.} We visualize the graphs generated by D2G2 and various baselines to qualitatively evaluate their dynamic graph generation capabilities. In the visualization, nodes of the temporal graphs span all temporal snapshots, while arcs are edges reflecting connectivities among the nodes of the temporal snapshots. Node attributes are represented by the colors (white ones denote those who are incapable of learning node features) of nodes. We use the protein dataset as the running example and more qualitative evaluations are available in the supplementary material. Figure~\ref{fig: topology} shows the graphs generated for the protein dataset by all the methods. It clear that D2G2 is superior in capturing the underlying characteristics of graphs and is closest to the ground truth in Figure~\ref{fig: topology}(a), in terms of both edges and nodes. GraphVAE is the second best one whose edge weight generation are worse than D2G2.

\begin{figure*}[!htb]
	\centering
	\subfigure[(a) Ground truth]{\includegraphics[width=0.24\textwidth]{./figures/all/real.png}\vspace{-0.4cm}}
	\subfigure[(b) $z$, $z'$ and $z''$ (without f)] {\includegraphics[width=0.24\textwidth]{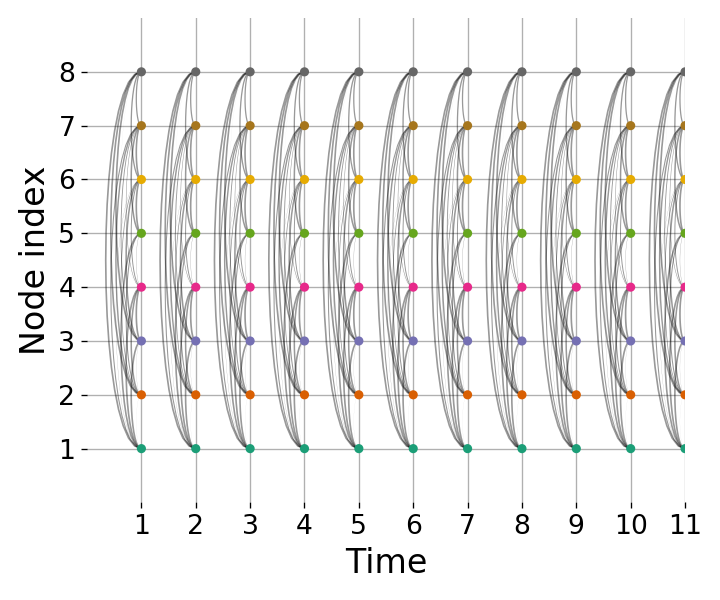}\vspace{-0.4cm}}
	\subfigure[(c) f and combined $z$, $z'$ and $z''$] {\includegraphics[width=0.24\textwidth] {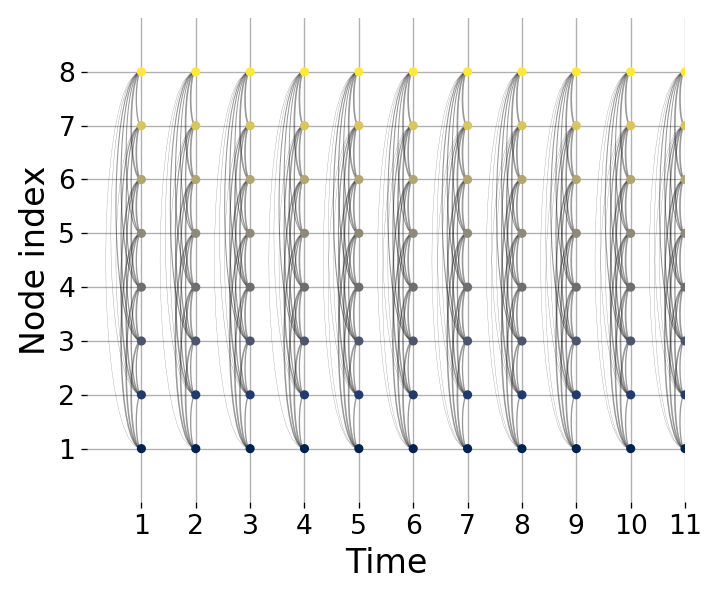}\vspace{-0.4cm}}
	\vspace{-0.3cm}
	\caption{Visualization of graphs from Protein dataset and generated by all ablated methods.\vspace{-0.5cm}}
	\label{fig: ablation}
\end{figure*}

\textbf{Evaluation of the generated disentangled patterns.}
We further qualitatively evaluate whether our D2G2 indeed disentangles dynamic and static patterns in nodes and edges, by randomly sampling one of $f$, $z$, $z'$ and $z''$ while keeping the other three fixed. This can control the respective factors of static patterns, edge dynamics, node dynamics as well as edge-node joint dynamics, respectively. We have observed numerous interesting patterns among different datasets and here list some of them as shown in Figure~\ref{fig: caseStudy}, while more are in the supplementary material.

The results demonstrate that the learned representation is indeed factorized and can control the corresponding patterns related to them. For example, as shown in Figure \ref{fig: caseStudy}(a), when we fix $z$, $z'$ and $z''$ while vary $f$, the whole dynamic graph changes evenly across different time snapshots, which verifies that $f$ indeed controls the invariant patterns of the dynamic graphs. Differently, when we instead vary $z$ by fixing $z'$, $z''$, and $f$, the topology in different time snapshots for each graph varies, which demonstrate that $z$ indeed effectively controls the non-stationary patterns in each dynamic graph; we can also see $z'$ is capable of finding time variant factor controlling attribute as the colors (i.e., the corresponding node attributes values they reflect) vary instead of the topology; when $z''$ is the only varying factor, node and edges change across different graphs in Figure \ref{fig: caseStudy}(d) as composed to either edges change in Figure \ref{fig: caseStudy}(b) or node changes in Figure \ref{fig: caseStudy}(c).

%

%
%
%

\textbf{Ablation Studies.} We also perform two ablations to confirm the necessity of the proposed factorization design. Without time-independent latent representation, the graphs generated in Figure~\ref{fig: ablation}(b) capture the underlying node characteristics but not topology. This might be because although the latent variables increase by respective dimension of ablated $f$, the dimension is still too small to fully capture its topology (yet intrinsic) characteristics. On the other hand, without separated topology and node latent variables, Figure~\ref{fig: ablation}(c) reveals that the generated topology looks close to ground truth but the node characteristics do not. This illustrates the necessity of separating topology and node representations. As otherwise the model might focus on the learning of more complex topology representation and ignore node representation, or vice versa.

\begin{figure}[!htb]\vspace{-0.25cm}
	\centering
	\includegraphics[width=0.35\textwidth]{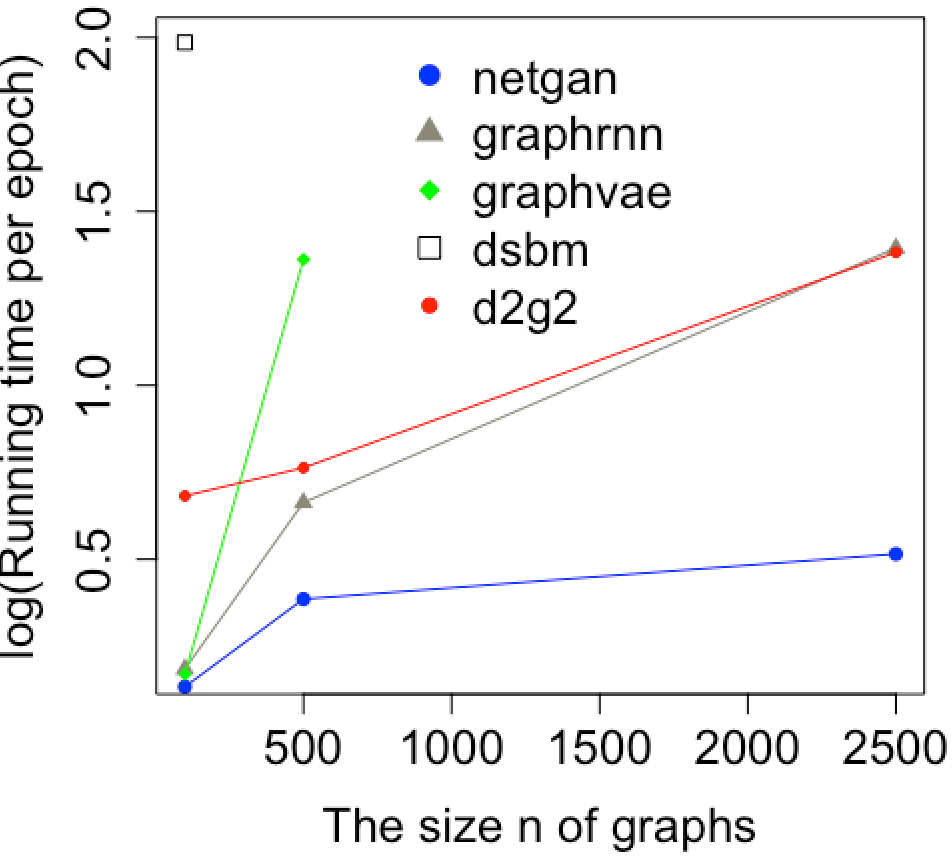}\vspace{-0.3cm}
	\caption{Scalability comparison of all the methods.\vspace{-0.3cm}}
	\label{fig:scalability}
\end{figure}

\textbf{Scalability.} We also illustrate and analyze the scalability in the number of nodes of our D2G2 method against comparison methods, as shown in Figure \ref{fig:scalability}, where the y-axis is the logarithm (base 10) of the runtime in seconds. As can be seen, our method achieves competitive scalability in general, which is similar to GraphRNN and better than GraphVAE and DSGM. NetGAN is known for advantageous scalability yet lower interpretability due to the utilization of random walks. GraphVAE and DSGM cannot run with affordable memory and time for graphs larger than 1000 nodes in our experiments.

\vspace{-0.3cm}

\section{Conclusions}
\label{sec: conclusion}

Generative models for real-world graphs have attracted significant attentions since it is an attractive way to learn high-level graph representations. However, most of existing graph generative models do not explicitly consider the dynamically evolving topology and attribute information. To fill the gap, this paper proposes a novel deep generative model, known as D2G2, for attributed temporal graphs. We consider the attributed temporal graph generation as a disentanglement problem and present a model consisting of two derived penalties for disentanglements between topology and attribute as well as between time-dependent and time-varying factors. The experimental evaluation results show the flexibility and versatility of D2G2 in generating dynamic graphs. One immediate future direction is to extend
these results in conjunction with our previous works~\cite{zhang2019faht,zhang2020online} for unbiased scene graph generation.


\section*{Acknowledgement}
This work was supported by the National Science Foundation (NSF) Grant No. 1755850, No. 1841520, No. 2007716, No. 2007976, No. 1942594, No. 1907805, a Jeffress Memorial Trust Award, Amazon Research Award, NVIDIA GPU Grant, and Design Knowledge Company (subcontract number: 10827.002.120.04).

\bibliographystyle{IEEEtran}
\vspace{-0.2cm}
\bibliography{typeinst}

%
%
%
%

\end{document}